\documentclass[lettersize,journal]{IEEEtran}
\usepackage{amsmath,amsfonts}
\usepackage{algorithmic}
\usepackage{algorithm}
\usepackage{array}

\usepackage[caption=false,font=small,labelfont=rm,textfont=rm]{subfig}
\usepackage{textcomp}
\usepackage{stfloats}
\usepackage{url}
\usepackage{verbatim}
\usepackage{graphicx}
\usepackage{cite}

\usepackage{booktabs}
\usepackage{multirow}
\usepackage{multicol}
\usepackage{tabularx}
\usepackage{float}
\usepackage{makecell}
\usepackage{enumitem}
\usepackage{amssymb}
\usepackage[colorlinks]{hyperref}
\usepackage{cleveref}
\usepackage{bm}
\usepackage{color}
\usepackage{soul}

\begin{document}

\title{Context-Enhanced Detector For Building Detection From Remote Sensing Images}

\author{Ziyue Huang, Mingming Zhang, Qingjie Liu,~\IEEEmembership{Member,~IEEE}, Wei Wang, Zhe Dong, and Yunhong~Wang,~\IEEEmembership{Fellow,~IEEE}
\thanks{Ziyue Huang, Mingming Zhang, Qingjie Liu, and Yunhong Wang are with the State Key Laboratory of Virtual Reality Technology and Systems, Beihang University, Beijing 100191, China, and also with the Hangzhou Innovation Institute, Beihang University, Hangzhou 310051, China (e-mail: ziyuehuang@buaa.edu.cn; sara\_@buaa.edu.cn; qingjie.liu@buaa.edu.cn; yhwang@buaa.edu.cn). 

Wei Wang, and Zhe Dong are with the National Disaster Reduction Center of China, Beijing 100124, China (e-mail: 732196152@qq.com; dz1227@foxmail.com).} 
}

\markboth{IEEE Transactions on Geoscience and Remote Sensing}%
{Shell \MakeLowercase{\textit{et al.}}: A Sample Article Using IEEEtran.cls for IEEE Journals}


\maketitle

\begin{abstract}
The field of building detection from remote sensing images has made significant progress, but faces challenges in achieving high-accuracy detection due to the diversity in building appearances and the complexity of vast scenes. 
To address these challenges, we propose a novel approach called Context-Enhanced Detector (CEDet). 
Our approach utilizes a three-stage cascade structure to enhance the extraction of contextual information and improve building detection accuracy. 
Specifically, we introduce two modules: the Semantic Guided Contextual Mining (SGCM) module, which aggregates multi-scale contexts and incorporates an attention mechanism to capture long-range interactions, and the Instance Context Mining Module (ICMM), which captures instance-level relationship context by constructing a spatial relationship graph and aggregating instance features. 
Additionally, we introduce a semantic segmentation loss based on pseudo-masks to guide contextual information extraction. Our method achieves state-of-the-art performance on three building detection benchmarks, including CNBuilding-9P, CNBuilding-23P, and SpaceNet.

\end{abstract}


\section{Introduction}
\IEEEPARstart{B}{uilding} detection plays a pivotal role in various remote sensing applications, including urban planning \cite{kamusoko2017importance}, earthquake disaster reduction \cite{APPEarthquakeYOLO}, and mapping \cite{APPMAPPING}. 
With the advent of Convolutional Neural Networks (CNNs) \cite{AlexNet, ResNet}, many studies \cite{li2018hough, MBFF, MTCNN} have been proposed to effectively detect buildings in remote sensing images.
In this paper, we represent buildings using oriented bounding boxes (OBBs) \cite{DOTA} and use the object detection method to extract buildings.

As crucial components of human activities, buildings showcase significant disparities in appearance, shape, structure, and material. 
Such extensive heterogeneity, combined with the complexity of environments, poses a significant challenge to the detector’s accuracy. 
To overcome the challenge, some studies \cite{Context_CADNet, Context_CFENet, Context_PCI, Building_BOMSCNet} incorporate contextual information in the detection process.  
These studies can be broadly categorized into multi-scale feature context modeling and relational context modeling \cite{Context_ISNet}. 
Multi-scale context feature modeling aims to capture contextual features by utilizing dilated convolution \cite{Context_ASPP} or pyramid pooling \cite{PSPNet} to expand the receptive fields of features. 
However, existing methods \cite{CATNet, A2FPN, FAFPN, wang2022bridging} lack direct long-range feature interactions \cite{NonLocal}, which limits their ability to comprehend the overall scene. 
Relational context modeling analyzes the spatial relationships between instances \cite{RelationNetwork, HTD, Relation_CLT, Context_PCI} to improve detection accuracy. 
However, these methods have certain limitations when applied to building detection. 
The complex environment of building groups can lead to numerous low-quality or erroneous region proposals, which undermine the validity of relationship reasoning.

To overcome the limitations of current context-enhancement methods, we propose Context-Enhanced Detector (CEDet). 
CEDet introduces the Semantic Guided Contextual Mining (SGCM) module to facilitate multi-scale feature context modeling. 
The SGCM incorporates a self-attention mechanism to strengthen long-range feature dependencies. 
Moreover, SGCM integrates multi-scale features to generate rich semantic features encompassing global context and utilizes a semantic segmentation loss based on pseudo-masks to guide contextual information extraction. 
To enhance the modeling of relational context, CEDet adopts a multi-stage framework inspired by Cascade R-CNN \cite{Cascade}, aiming to improve the quality of region proposals. 
Finally, CEDet introduces a Context Enhancement OR-CNN Head (CE Head), which separates the classification and regression tasks. 
CE Head incorporates the Instance Context Mining Module (ICMM) to acquire spatial relational contextual information. 
This integration significantly enhances the ability to identify buildings within intricate scenes. 
Our contributions could be summarized as follows:
\begin{enumerate}
    \item We present Context-Enhanced Detector (CEDet), a novel approach for building detection in remote sensing images. 
    CEDet incorporates a cascade structure to enhance the accuracy of the detection process. 
    \item To enhance the multi-scale feature context, we propose the Semantic Guided Contextual Mining (SGCM) module. 
    The SGCM utilizes a self-attention mechanism to strengthen long-range feature dependencies. 
    Additionally, a semantic segmentation loss based on pseudo-masks is introduced to guide the extraction of contextual information. 
    \item The Instance Context Mining Module (ICMM) is introduced to capture contextual information between instances by leveraging spatial relationships. This module significantly improves the detector's ability to identify buildings in complex environments.
    \item Our CEDet achieves state-of-the-art (SOTA) performance on the CNBuilding-9P, CNBuilding-23P, and SpaceNet datasets. 
    Compared with the baseline OR-CNN \cite{ORCNN}, CEDet demonstrates improvements of 2.1\% in AP50 on CNBuilding-9P, 2.3\% on CNBuilding-23P, and 2.4\% on SpaceNet.

\end{enumerate}

\section{Related Work}
\subsection{Oriented Object Detection}
Oriented object detection has received extensive attention in remote sensing image understanding \cite{ROITrans, S2ANet, R3Det, KFIoU, KLD, ORCNN}, since the OBB representations can finely capture the appearance of objects in remote sensing images \cite{DOTA}. 
Ding et al. \cite{ROITrans} proposed RoI-Transformer, which introduced rotated RoIAlign to extract spatially invariant features. 
Xu et al. \cite{GlidingVertex} proposed Gliding Vertex, which employed the offset relative to the circumscribed rectangle to represent oriented objects. 
Yang et al. \cite{R3Det} proposed R3Det, utilizing the feature alignment module to address the problem of feature inconsistency in oriented detection. 
Han et al. \cite{ReDet} utilized an invariant rotation network and rotation covariant head to enhance the quality of rotation feature representation. 
Qin et al. \cite{MRDet} introduced an anchor-free oriented object detection model, achieving a balance between speed and accuracy. 
Han et al. \cite{S2ANet} employed rigid deformable convolution \cite{DCN} to achieve feature alignment for single-stage detectors. 
Several studies, including GWD \cite{GWD}, CSL \cite{CSL}, KFIoU \cite{KFIoU}, and KLD \cite{KLD}, have explored effective regression loss for oriented object detection.
These methods aimed to address the challenges of discontinuity and aliasing in OBB representation.
Li et al. \cite{OrientedReppoints} improved Reppoints \cite{Reppoints} to enable oriented detection. 
Xie et al. \cite{ORCNN} proposed oriented R-CNN by utilizing oriented RPN to generate high-quality rotation proposals, simplifying the pipeline of the two-stage detector. 
Experimental results demonstrate that oriented R-CNN achieves a favorable balance between speed, accuracy, and simplicity.
Hence, we adopt oriented R-CNN as our baseline.

\subsection{Building Segmentation}
Building extraction has been the focus of numerous methods due to the potential value of building information in various applications. 
Semantic segmentation serves as a typical task setting for building extraction. 
Alshehhi et al. \cite{alshehhi2017simultaneous} proposed a patch-based CNN architecture for building segmentation. 
Hamaguchi et al. \cite{MTCNN} presented a multi-task U-Net network that incorporates contextual information through road extraction tasks. 
A Multi-scale learning strategy is also adopted to address the problem of scale variation. 
Yang et al. \cite{yang2018building} designed a dense-attention network based on DenseNet \cite{DenseNet}.  
They combined a spatial attention mechanism with DenseNet to fuse features at different scales dynamically. 
Griffiths et al. \cite{griffiths2019improving} integrated multi-source data (optical, lidar, and building footprint) and employed Morphological Geodesic Active Contours to enhance the quality of annotations. 
Wei et al. \cite{wei2019toward} proposed an improved fully convolutional network to obtain building footprint masks, which were then transformed into polygons using a polygon regularization algorithm. 
Zhu et al. \cite{zhu2021adaptive} presented an adaptive polygon generation algorithm (APGA) that generates sequences of building vertices and arranges them to form polygons. 
Hosseinpour et al. \cite{CMGFNet} proposed CMGFNet, which incorporates digital surface model features with conventional optical features to obtain robust features. 
Visual transformer (ViT) \cite{ViT} has demonstrated effective modeling of global dependencies. 
Based on this, Wang et al. \cite{BuildFormer} introduced BuildFormer, a method that leverages ViT to extract vital global features for finer building segmentation.
Zorzi et al. \cite{zorzi2022polyworld} utilized Graph Neural Network (GNN) to propagate global information to vertices, enhancing detection accuracy. 
While building extraction based on semantic segmentation has made significant progress, it is essential to note that the masks predicted by semantic segmentation models may not effectively extract individual building instances. 
This limitation restricts the applicability of these methods in real-world scenarios.

\subsection{Building Detection}
Building detection is more challenging than building segmentation, but it offers valuable instance-level information, which is crucial for downstream applications. 
PolyRNN \cite{PolyRNN} and PolyRNN++ \cite{PolyRNN++} have approached instance-level segmentation as a contour polygon prediction task, inspiring subsequent building detection research. 
Li et al. \cite{li2019topological} introduced PolyMapper, a serialization method that converts the graph structure of roads into a sequential structure, enabling simultaneous extraction of building footprints and roads.
Xu et al. \cite{xu2021gated} proposed a centroid-aware head based on Mask R-CNN \cite{he2017mask}.
Li et al. \cite{li2021deep} developed a hybrid model for building polygon extraction, utilizing multiple networks to obtain bounding boxes, segmentation masks, and corners of buildings, and employing Delaunay triangulation to construct building polygons. 
Liu et al. \cite{HTCBuilding} improved HTC \cite{HTC} with a more robust backbone and dynamic anchor strategy to achieve high-quality building extraction. 
Hu et al. \cite{hu2023polybuilding} proposed a transformer-based polygonal building extraction model using a two-phase strategy for multi-task training.
Zhao et al. \cite{zhao2022rotation} integrated a rotated detector into building detection, introducing RotSegNet. 
Rotation detection enables the extraction of rotation-invariant features and effective detection of buildings in dense areas.
Compared to the segmentation methods, we are oriented to solve the problem of building detection in complex scenes, which can be further combined with the segmentation methods. 
\begin{figure*}[!htb]
	\centering
	\includegraphics[width=1\linewidth]{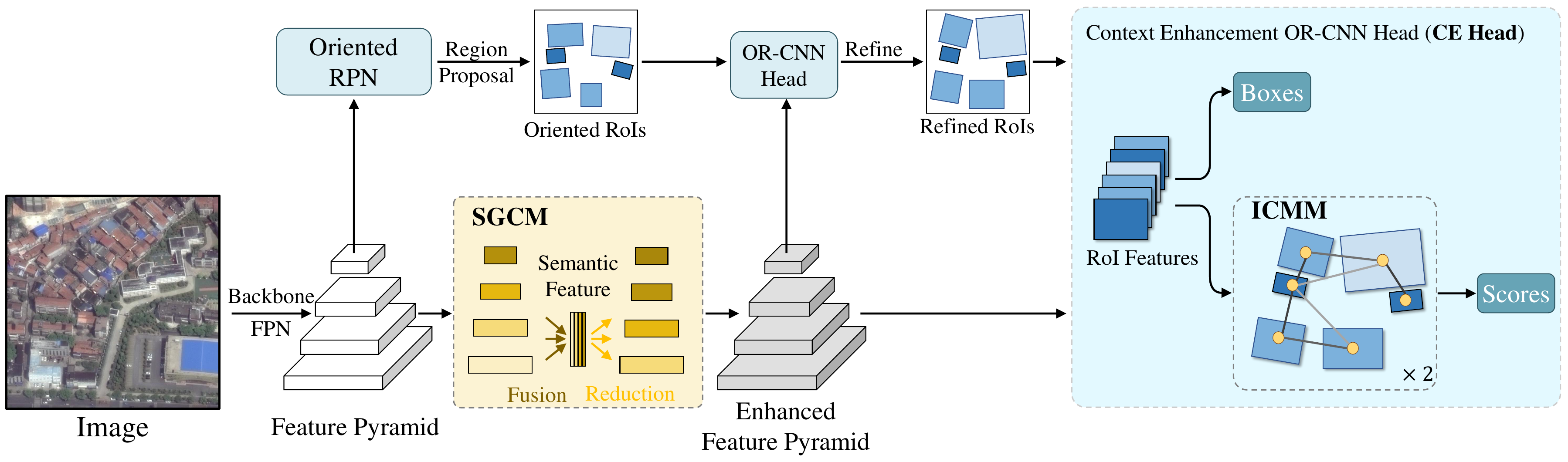}
	
	\caption{
Context-Enhanced Detector (CEDet) is a three-stage detection model for high-accuracy building detection. 
Semantic Guided Context Mining (SGCM) module can enhance the multi-scale feature context. 
Context Enhancement OR-CNN Head (CE Head) adopts the decoupling structure and obtains the relationship contextual information by  Instance Context Mining Module (ICMM). }
\label{fig:Architexture}
\end{figure*}

\subsection{Context-Enhancement Method} 
The incorporation of contextual information has been proven to be effective in improving the performance of detection \cite{RelationNetwork, HTD, Relation_CLT, Context_CADNet, Context_PCI, wang2022bridging} and segmentation \cite{Context_ASPP, Context_ISNet, A2FPN, CATNet, Building_BOMSCNet, Relation_HSDN, Context_CFENet}. 
For instance, Chen et al. \cite{Context_ASPP} proposed the ASPP module, which employs multiple parallel atrous convolutions to capture long-distance context. 
Jin et al. \cite{Context_ISNet} introduced ISNet, which incorporates semantic-level and image-level contextual information to augment the features. 
Both $A^2$-FPN \cite{A2FPN} and CATNet \cite{CATNet} proposed complex multi-scale feature interaction modules and employed global context aggregation to model global dependency. 
Zhou et al. \cite{Building_BOMSCNet} enhanced the network's ability to reason about contextual information by incorporating multi-scale features and graph reasoning.
Zheng et al. \cite{Relation_HSDN} focused on modeling semantic relationships between pixels and proposed the HSDN, utilizing these semantic relations for semantic decoupling and feature enhancement. 
Chen et al. \cite{Context_CFENet} explicitly established contextual relationships between low-level and high-level features.

Context modeling between instances is more common in detection tasks because the region features used by the two-stage detectors \cite{Faster-RCNN, ORCNN} naturally correspond to instance features. 
Hu et al. \cite{RelationNetwork} proposed an attention mechanism to automatically model the semantic and spatial relationships between regional features, enabling effective context modeling. 
Li et al. \cite{HTD} introduced a task decoupling detection framework that leverages local spatial aggregation and global semantic interactions to capture relational context. 
CLT-Det \cite{Relation_CLT} introduced a correlation transformer module to enhance the information interaction between regional features, improving contextual understanding. 
CAD-Net \cite{Context_CADNet} utilized multi-scale regional features to obtain contextual information, allowing the network to focus on scale-specific features for improved performance. 
PCI \cite{Context_PCI} established dynamic proposal graphs for refining classification and regression, incorporating contextual information into the detection process. 
Nevertheless, it is imperative to acknowledge that within intricate remote sensing scenes, noise poses challenges in achieving effective relationship modeling, consequently influencing detection performance. 

\begin{figure*}[htpb]
    \centering
    \includegraphics[width=0.9\linewidth]{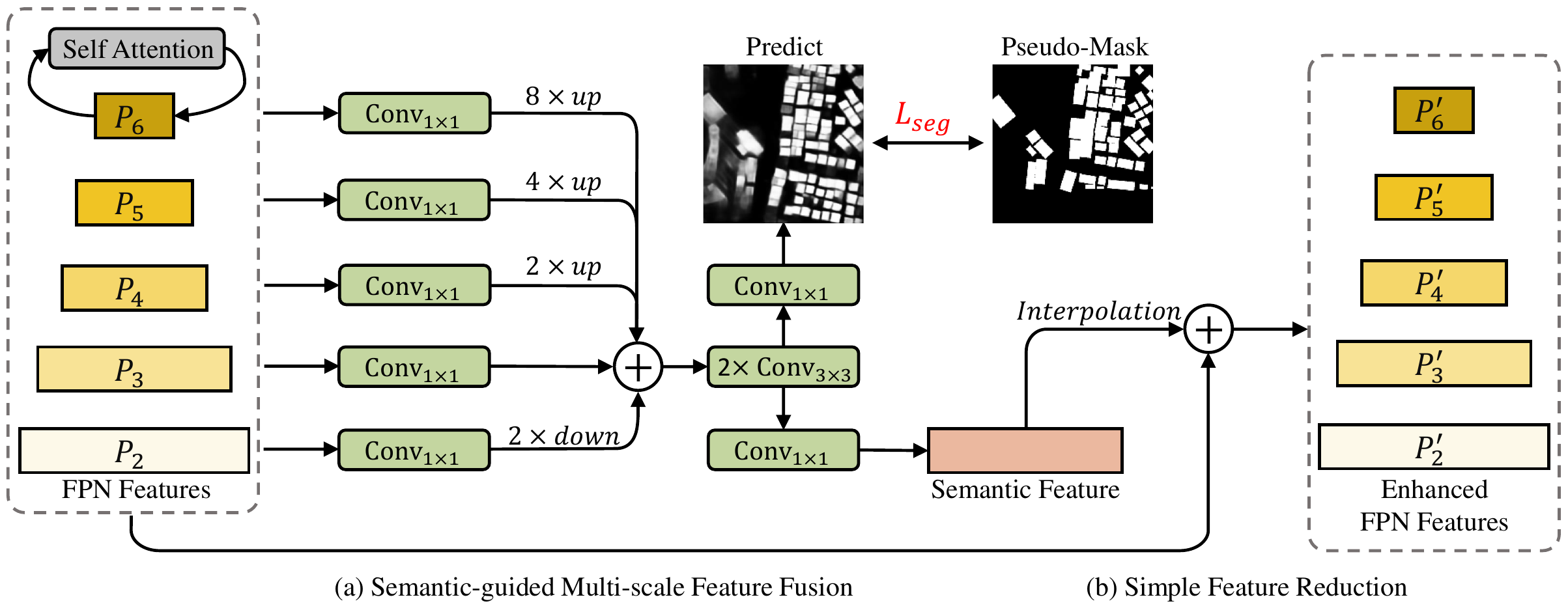}
    \caption{Semantic Guided Context Mining (SGCM) module has two phrases:  
    (a): Fusion, which enhances contextual information, performs feature fusion and uses semantic loss for supervision. 
    (b): Reduction, which enhances FPN features with the fused feature. 
    }
    \label{fig:SGCM}
\end{figure*}

\section{Method}
As shown in Fig. \ref{fig:Architexture}, we propose a Context-Enhanced Detector (CEDet), a three-stage building detection framework, for high-accuracy building detection.
Firstly, CEDet introduces a Semantic Guided Context Mining (SGCM) module to extract contextual information at the feature level. 
SGCM adopts semantic-guided multi-scale feature fusion to enhance multi-scale features of Feature Pyramid Network (FPN) \cite{FPN}. 
Subsequently, oriented Region of Interests (oriented RoIs) extracted by oriented Region Proposal Network (oriented RPN) are refined with enhanced features. 
Additionally, a Context Enhancement OR-CNN Head (CE Head) is designed for building classification and oriented bounding box regression. In building classification branch, Instance Context Mining Module (ICMM) is introduced to capture relational contextual information at the instance level. Besides, offsets predicted by oriented bounding box regression branch will improve building detection precision. 
We will detail the main components of CEDet in the subsequent subsections.

\subsection{Semantic Guided Context Mining}
Semantic Guided Context Mining (SGCM) module is proposed to extract contextual information by capturing the long-range dependence of features from FPN, including semantic-guided multi-scale feature fusion and simple feature reduction as shown in Fig. \ref{fig:SGCM}. 

Given multi-scale features $\{P_2, P_3, P_4, P_5, P_6\}$ from FPN, a self-attention block \cite{Transformer} is firstly adopted to capture the long-range dependence from $P_6$ by self-attention operation.
Then, features $\{P_2, P_3, P_4, P_5, P_6\}$ go through independent $1\times1$ convolutional layers and are scaled to the same size with $P_3$ by bilinear interpolation. 
Subsequently, the multi-scale fused feature is obtained by adding scaled features at all levels, in which each location establishes an implicit global relationship with the whole image.

After the multi-scale fused feature undergoes two $3\times3$ convolutional layers, a single $1\times1$ convolutional layer is employed to acquire the enriched semantic feature. 
Another $1\times1$ convolutional layer is utilized to generate pseudo-masks for subsequent segmentation loss computation. 
Ultimately, the enhanced multi-scale features are attained through the element-wise addition of the rich semantic feature to the original multi-scale features extracted from FPN. 
This feature reduction process incorporates valuable semantic and multi-scale contextual information into the multi-scale features obtained from FPN and diminishes the semantic gap \cite{LibraRCNN} among these features. 

To effectively extract contextual information, the fused feature within the proposed SGCM is supervised by a semantic segmentation loss $L_{seg}$. 
Since ground truth masks are unavailable, we use OBB annotations to generate pseudo-masks that serve as surrogate ground truth for calculating the semantic segmentation loss. 
Our experiments demonstrate that these pseudo-masks strongly resemble the actual ground truth masks in many scenes, primarily due to the prevalent rectangular shape of most buildings. 
Consequently, despite the lack of genuine ground truth, the predicted pseudo-masks effectively depict the outlines of buildings. 
Finally, the semantic segmentation loss is calculated by pixel-wise cross-entropy loss: 

\begin{equation}
L_{Seg} = CrossEntropy(M^*, M_{pseudo})
\label{eq:L_seg}
\end{equation}
where $M^*$ denotes the predicted mask, $M_{pseudo}$ denotes the pseudo-mask, $L_{Seg}$ is the semantic segmentation loss.

\subsection{Instance Context Mining Module}
Spatial relationships \cite{RelationNetwork, HTD, Relation_CLT} among building instances can help to identify buildings more accurately from complex backgrounds. 
Unlike contextual information extraction at the feature level, instance-level relationship modeling adopts radial symmetric feature aggregation to bring stronger spatial invariance. 
Moreover, the introduction of RRoIAlign \cite{ROITrans} and the deepening of semantic level reduce the interference of noise, which is helpful to the reasoning of high-order relationships. 

However, there are two problems with directly using oriented RoIs generated by oriented RPN for relational reasoning. 
Firstly, the presence of redundant RoIs introduces instability to the feature distribution of relational context, thereby increasing the difficulty of relational reasoning. 
Secondly, RoIs may contain multiple background regions, such as clusters of buildings, fragmented parts of buildings, and false detections. 
These background regions further hinder the extraction of meaningful relational context features. 
Therefore, we propose the Instance Context Mining Module (ICMM), which employs a feature aggregation method to extract spatial relational context and address the problems. 

\begin{figure}[!htb]
	\centering
	\includegraphics[width=1\linewidth]{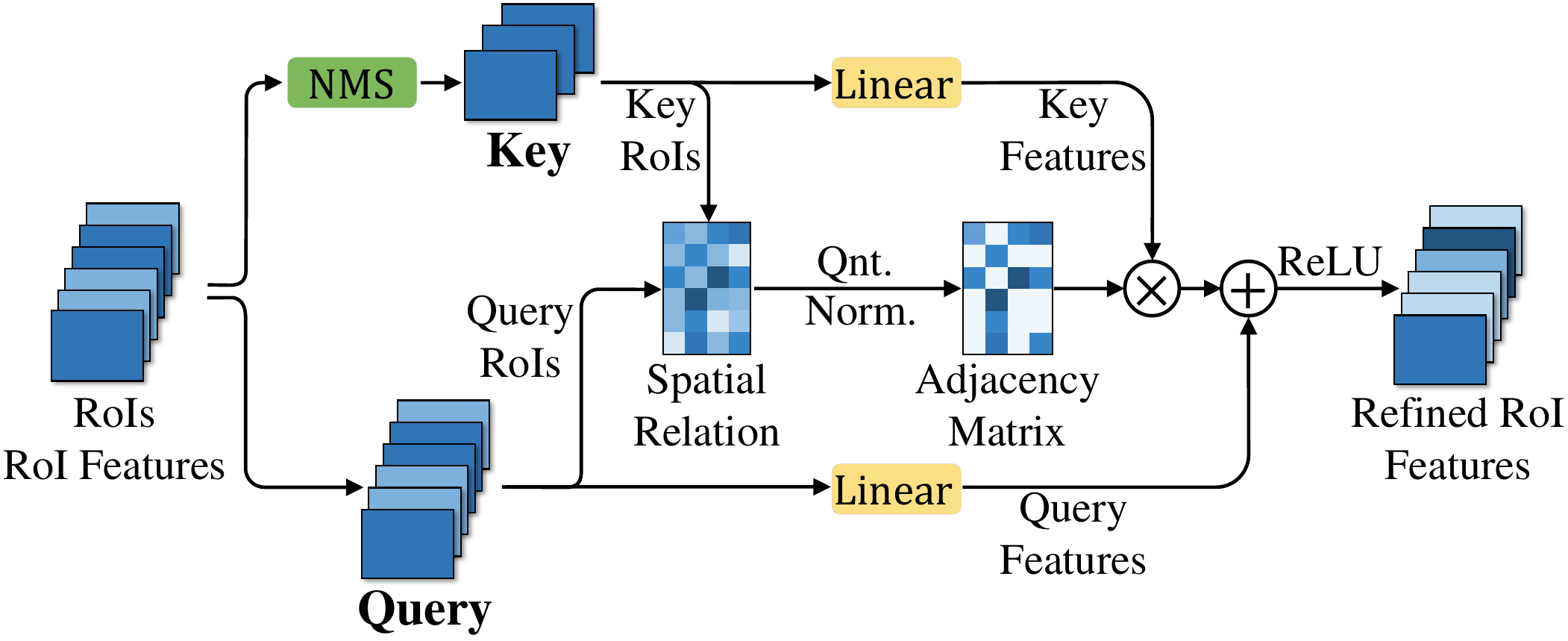}
	
	\caption{Instance Context Mining Module (ICMM) uses spatial relationships between RoIs to extract instance-level contextual features. }
\label{fig:ICMM}
\end{figure}

To achieve more efficient relational modeling, we use Non-maximum suppression (NMS) to suppress the duplicate RoIs and the false RoIs. 
As shown in Fig. \ref{fig:ICMM}, both  RoIs $B \in\mathbb{R}^{N\times 5}$ and the corresponding RoI features $F \in\mathbb{R}^{N\times C}$ are fed into the ICMM, where $N$ is the number of RoIs. 
Consistent with OR-CNN \cite{ORCNN}, RoI features $F$ are cropped from FPN features by using $7\times7$ size RRoIAlign and mapped to $C=1024$ dimensions by a flattening operation and two fully connected layers with ReLU. 

Each RoI contains five elements $(x, y, w, h, \theta)$, where $(x, y)$ is the center coordinates, $(w, h)$ is the width and height, and $\theta$ is the rotation angle of the RoI. 
Unlike standard GCN \cite{GCN}, RoIs are split into queries and keys for separate processing, which allows for better handling of contextual features. 
Query branch maintains all RoIs, termed as $B^q \in\mathbb{R}^{N\times 5}$. 
The corresponding features are mapped through a linear layer to obtain query features $F^q \in \mathbb{R}^{N\times C}$. 
The key branch uses non-maximum suppression (NMS) with an Intersection over Union (IoU) threshold of 0.5 to filter noises and obtain key RoIs $B^k \in\mathbb{R}^{M\times 5}$, where $M$ is the number of RoIs after NMS. 
The corresponding features are mapped through an independent linear layer to obtain key features $F^k \in\mathbb{R}^{M\times C}$. 

The normalized center distance is used to model building instance spatial relationships. 
Let the $i$-th RoI in $B^q$ be $b_i^q=(x_i^q, y_i^q, w_i^q, h_i^q, {\theta}_i^q)$, and the $j$-th RoI in $B^k$ be $b_j^k=(x_j^k, y_j^k, w_j^k, h_j^k, {\theta}_j^k)$, then the spatial relationship between $b_i^q$ and $b_j^k$ can be obtained by the exponential transformation of center distances:  
\begin{equation}
\Delta_{i,j} = \left(\frac{(x_i^q - x_k^k)}{w_i^q}, \frac{(y_i^q - y_j^k)}{h_i^q} \right)
\label{eq:Spatial}
\end{equation}
\begin{equation}
S_{i,j} = S(b_i^q, b_j^k) = \exp\{-\frac{||\Delta_{i,j}||_2}{2}\}
\label{eq:Spatial2}
\end{equation}
where $||\cdot||_2$ is the $L_2$ norm, $S=[S_{i,j}] \in \mathbb{R}^{N\times M}$ is the spatial relation matrix between queries and keys. 
The minimum value of $w_i^q$ and $h_i^q$ is limited to 56 to avoid the normalized center distance between small RoIs and other RoIs being too large to establish an effective spatial relationship. 

Then, a fixed threshold $t$ is used to quantify the spatial relation matrix: 
\begin{equation}
A_{i,j} = \left\{
\begin{aligned}
0 &,\quad S_{i,j} < t \\
1 &,\quad S_{i,j} \geq t
\end{aligned}
\right.
\label{eq:Quantify}
\end{equation}
where $t$ empirically sets to 0.1. 
$A=[A_{i,j}] \in \mathbb{R}^{N\times M}$ is the adjacency matrix describing spatial relations. 
After that, the discrete adjacency matrix is normalized by degree matrix $D\in \mathbb{R}^{N\times N}$ to obtain the final adjacency matrix $\hat{A}$: 
\begin{equation}
D_{i,j} = \left\{
\begin{aligned}
\sum_{k}{A_{i,k}} & ,\quad i=j \\
0 \quad  & ,\quad i \neq j
\end{aligned}
\right.
\label{eq:Quantify}
\end{equation}

\begin{equation}
\hat{A} = D^{-1}A
\label{eq:Norm2}
\end{equation}

Then key features $F^k$ are multiplied by normalized adjacency matrix $\hat{A}$ to obtain context-enhanced features $F' \in\mathbb{R}^{N\times C}$. 
Query features are added to preserve the original semantic information: 
\begin{equation}
F' = \mathrm{ReLU}(\hat{A} F^k + F^q)
\label{eq:ContextFusion}
\end{equation}
where $\mathrm{ReLU}$ denotes the ReLU activate function. 
ICMM can be repeated multiple times to capture rich instance relationships, and our experiments show that two ICMMs achieve a trade-off between performance and efficiency. 
Finally, the ICMM-enhanced RoI features will pass through two fully connected layers for building classification. 

\subsection{Detection}
This subsection will introduce the overview of the proposed CEDet, as well as the training and inference processes. 
Using multi-scale features obtained from FPN, the oriented RPN predicts the offsets and classification scores of anchors on each scale. 
We define the loss functions of RPN as $L_{RPNCls}$ and $L_{RPNReg}$, which is consistent with the OR-CNN \cite{ORCNN}. 
Then, the oriented RPN outputs 2,000 oriented proposals as oriented RoIs. 
SGCM module enhances multi-scale features and is supervised by the semantic segmentation loss $L_{Seg}$ that is calculated as defined in Eqs. \ref{eq:L_seg}. 

Then, based on enhanced multi-scale features of FPN, an OR-CNN Head refines oriented RoIs by predicting the offset and classification of oriented RoIs, which is supervised by the classification loss $L_{H_1 Cls}$ and the regression loss $L_{H_1 Reg}$. 
Finally, refined oriented RoIs $B_1 \in \mathbb{R}^{2000\times 5} $ and the corresponding classification scores $C_1 \in \mathbb{R}^{2000\times 2}$ can be obtained through the offsets and scores predicted by OR-CNN Head. 
As a binary classification for building detection task, each classification score has two dimensions, which represent the probability of background and building. 

The training process of CE Head is the same as that of OR-CNN Head, despite using ICMM to capture the relationships.  
CE Head calculates the classification loss $L_{H_2 Cls}$ and the regression loss $L_{H_2 Reg}$, and obtains the final bounding boxes $B_2 \in \mathbb{R}^{2000\times 5} $ and classification score $C_2 \in \mathbb{R}^{2000\times 2} $. 
The total loss during training is:  
\begin{equation}
\begin{split}
L_{Total} = L_{RPNCls} + L_{RPNReg} + L_{Seg}\\
 + L_{H_1 Cls} + L_{H_1 Reg}\\
 + L_{H_2 Cls} + L_{H_2 Reg}
\label{eq:total_loss}
\end{split}
\end{equation}
In the inference stage, a series of predicted bounding boxes and corresponding classification scores are obtained through the same process. 
We take the final bounding boxes $B_2$ as the output. 
Consistent with Cascade R-CNN \cite{Cascade}, the average scores $C_{avg} = (C_1 + C_2) / 2$ are used as the final classification scores.

 The NMS is used as a post-processing to handle redundant detection boxes, with the threshold set to 0.1. 
 Due to the high density of buildings, we keep a maximum of 300 boxes per image. 

\section{Experiments}

\begin{figure*}[!htb]
	\centering
	\includegraphics[width=1\linewidth]{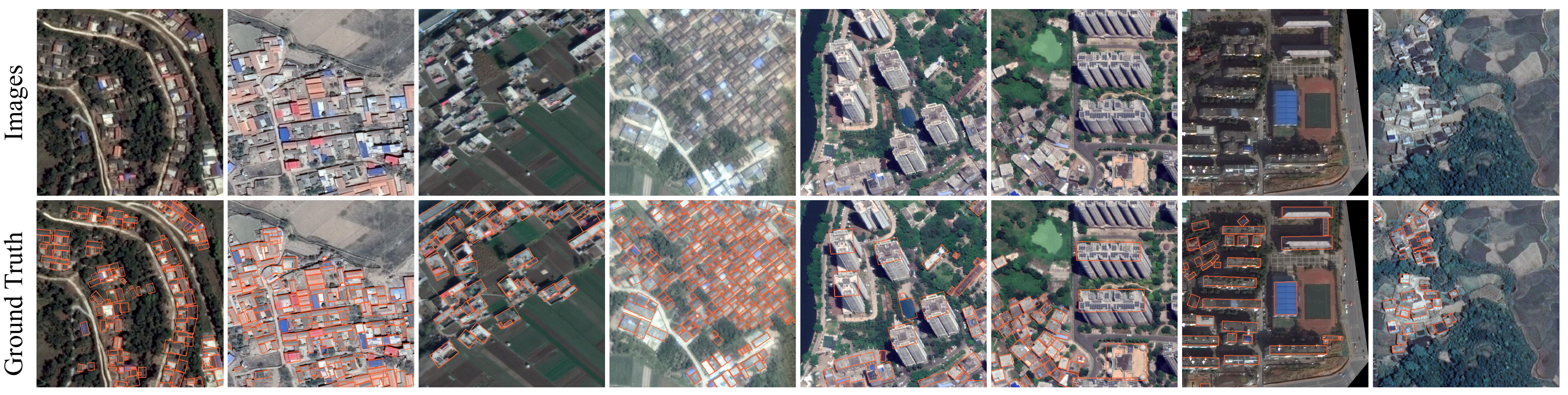}
	
	\caption{
 Some examples in the CNBuilding-9P. 
 CNBuilding-9P dataset covers nine provinces in China and contains 50,782 images with 1,210,968 building instances. 
 The first row is the original image, and the orange boxes in the second row represent the ground truth. 
 Zoom in to see more details. }
\label{fig:CNBuilding_exapmle}
\end{figure*}

\subsection{Dataset and Metrics}
We perform experiments on three datasets:
(1) CNBuilding-9P is a challenging large-scale building dataset covering nine provinces in China: Gansu (GS), Guangdong (GD), Guangxi (GX), Hainan (HI), Hunan (HN), Jilin (JL), Shandong (SD), Sichuan (SC), and Zhejiang (ZJ). 
This dataset encompasses a multitude of intricate scenes, including urban areas, countryside landscapes, farmland, forested areas, and mountainous regions, thereby encompassing the entirety of conceivable building types, such as residential housing, factories, shopping centres, warehouses, stadiums, and more. 
Images in the CNBuilding-9P dataset are collected from GoogleEarth with 50,782 images and 1,210,968 buildings that vary significantly in size, structure, and appearance. 
Images in CNBuilding-9P are manually labeled with instance-level OBB annotations, and some examples are shown in Fig. \ref{fig:CNBuilding_exapmle}. 
Detailed information regarding the images and building instances for each province's training and test sets can be found in Table \ref{tab:CNData-9P}. 
The validation set is derived by randomly sampling one-tenth of the data from the training set. 
Subsequently, all comparison methods, including our proposed CEDet, are trained using the training and validation sets, and the resulting detection performances on the test set are reported. 
All ablation studies are exclusively conducted using CNBuilding-9P.
\begin{table}[htbp]
  \centering
  \caption{Number of images and buildings in CNBuilding-9P dataset}
      \setlength{\tabcolsep}{3.2mm}{
    \begin{tabular}{c|cc|cc}
    \toprule[1pt]
    \multirow{2}[1]{*}{Province} & \multicolumn{2}{c|}{Train} & \multicolumn{2}{c}{Test} \\
          & \# Images & \# Instances & \# Images & \# Instances \\
    \midrule
    \midrule
    GS    & 5,809  & 87,260  & 600   & 11,110  \\
    GD    & 4,871  & 103,377  & 654   & 18,548  \\
    GX    & 4,328  & 71,969  & 522   & 10,951  \\
    HI    & 4,840  & 101,884  & 665   & 18,278  \\
    HN    & 4,226  & 111,709  & 674   & 16,041  \\
    JL    & 5,109  & 101,046  & 684   & 13,939  \\
    SD    & 5,992  & 150,792  & 700   & 27,501  \\
    SC    & 4,732  & 208,808  & 638   & 27,785  \\
    ZJ    & 5,078  & 111,958  & 660   & 18,012  \\
    \midrule
    \midrule
    Total & 44,985  & 1,048,803  & 5,797  & 162,165  \\
    \bottomrule[1pt]
    \end{tabular}%
    }
  \label{tab:CNData-9P}%
\end{table}%

	

\begin{figure}[!htb]
	\centering
	\includegraphics[width=1.0\linewidth]{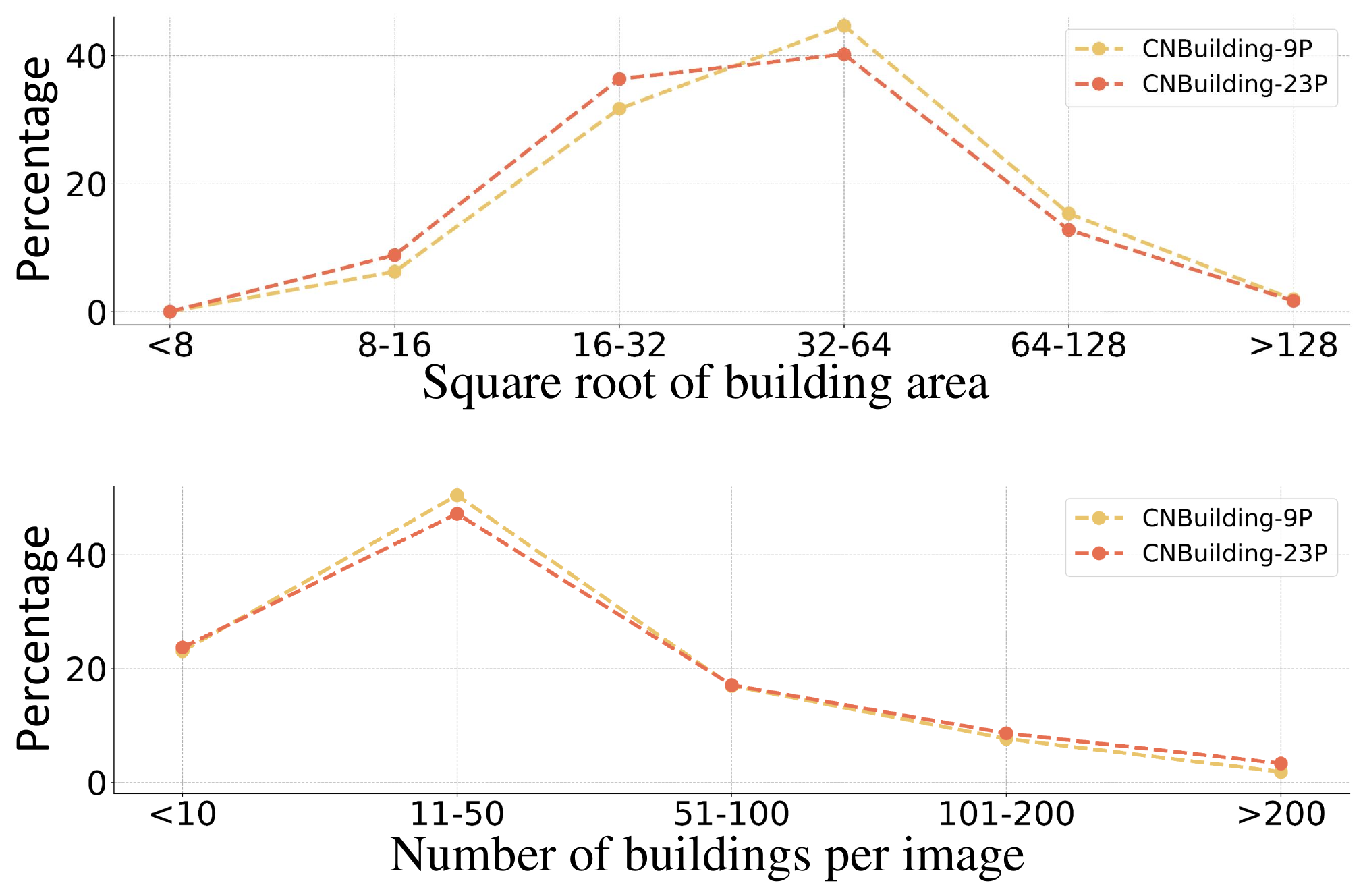}
	
	\caption{The statistic of CNBuilding-9P and CNBuilding-23P, including the square root of building area and the number of buildings per images.}
\label{fig:CNBuilding_Area}
\end{figure}

(2) CNBuilding-23P serves as an extension of the CNBuilding-9P dataset, encompassing an additional 14 areas: Chongqing (CQ), Fujian (FJ), Guizhou (GZ), Hebei (HE), Heilongjiang (HL), Hubei (HB), Jiangxi (JX), Liaoning (LN), Inner Mongolia (NM), Qinghai (QH), Shanxi (SX), Yunnan (YN), Tibet (XZ), and Other-Provinces (OP). 
With a total of 139,217 images and 3,623,425 buildings, CNBuilding-23P represents a scale almost three times larger than that of CNBuilding-9P.
Figure \ref{fig:CNBuilding_Area} compares the square root distribution of building areas between CNBuilding-23P and CNBuilding-9P. 
While the area distributions and building densities of the two datasets exhibit similarities, CNBuilding-23P presents a greater number of small building instances, thereby introducing additional challenges. 
While the two datasets' area distributions and building densities exhibit similarities, CNBuilding-23P presents a greater number of small building instances, thereby introducing additional challenges. 
Due to the substantial scale of the CNBuilding-23P dataset, the experiment only reports the overall results obtained from this dataset.

(3) SpaceNet: SpaceNet is a public building dataset obtained from the DigitalGlobe WorldView 3 satellite \cite{van2018spacenet}. 
We select Paris, Shanghai, and Khartoum for evaluation in our experiments. 
These areas contain 633 images with 16,853 building instances, 3,351 images with 71,294 building instances, and 923 images with 25,371 building instances, respectively. 
The polygon annotations in the dataset are transformed into rotated box representations using the minimum bounding rectangle algorithm. 
Afterward, the dataset is randomly split into training, testing, and validation subsets, with a ratio of 3:1:1, and the detection performance on the test set is reported.

\begin{table*}[htbp]
  \centering

  \caption{Performance comparison on CNBuilding-9P dataset in terms of AP50 on each province, overall AP50 and AP75, and the inference speed (Frame Per Second, FPS). 
  The inference speed is tested on a single 2080TI, including post-processing.}
  \resizebox{2.0\columnwidth}{!}{
    \begin{tabular}{c|cccccccccccccc}
    \toprule[1pt]
    Stage & Method & GS    & GD    & GX    & HI    & HN    & JL    & SD    & SC    & ZJ   & AP50  & AP75  & FPS   & Year \\
    \midrule
    \multicolumn{1}{c|}{\multirow{3}[2]{*}{Single}} & Gliding Vertex \cite{GlidingVertex} & 55.5  & 51.1  & 62.0  & 61.0  & 63.8  & 70.8  & 58.2  & 39.3  & 55.6  & 55.8  & 25.9  & 10.5  & 2020 \\
          & KLD \cite{KLD}   & 53.3  & 51.5  & 63.0  & 61.1  & 64.0  & 70.9  & 57.6  & 41.2  & 55.9  & 56.0  & 30.0  & 9.3   & 2021 \\
          & KFIoU \cite{KFIoU} & 55.5  & 52.2  & 62.9  & 61.6  & 65.1  & 73.0  & 58.8  & 39.5  & 55.1  & 56.4  & 27.7  & 14.4  & 2022 \\
    \midrule
    \multicolumn{1}{c|}{\multirow{5}[4]{*}{Multi}} & FR-O \cite{Faster-RCNN}  & 56.5  & 52.1  & 62.8  & 62.2  & 64.9  & 72.0  & 58.9  & 40.6  & 56.9  & 56.8  & 29.2  & \textbf{19.7} & 2017 \\
          & RoI-Transformer \cite{ROITrans} & 56.7  & 51.8  & 62.3  & 61.9  & 65.0  & 72.4  & 58.7  & 40.2  & 55.7  & 56.6  & 32.2  & 19.1  & 2019 \\
          & OR-CNN \cite{ORCNN} & 56.0  & 52.0  & 62.9  & 62.0  & 64.6  & 72.4  & 58.6  & 40.1  & 55.1  & 56.4  & 31.5  & 18.6  & 2021 \\
          & CR-O \cite{Cascade} & 55.2  & 51.3  & 61.1  & 61.8  & 64.6  & 71.1  & 56.9  & 39.5  & 54.8  & 55.6  & 33.9  & 11.5  &  -\\
          & CEDet (Ours)   & \textbf{57.1} & \textbf{53.9} & \textbf{64.0} & \textbf{63.8} & \textbf{66.3} & \textbf{74.4} & \textbf{61.3} & \textbf{42.8} & \textbf{57.8} & \textbf{58.5} & \textbf{34.1} & 8.9   &  -\\
    \bottomrule[1pt]
    \end{tabular}%
}
  \label{tab:CN9P_Results}%
\end{table*}%

\subsection{Implementation Details}
All experiments are conducted using the MMRotate \cite{MMRotate} framework. 
As for the backbone network, ResNet50 \cite{ResNet} is employed and initialized with pre-trained parameters from ImageNet \cite{ImageNet} to match the default configuration in MMRotate \cite{MMRotate}.
During training, images are resized to $800\times800$ using bilinear interpolation, and random horizontal flips with a probability of 0.5 are applied for data augmentation. 
Normalization is performed on the images using mean and variance values obtained from the ImageNet dataset. 
For image pre-processing in the test phase, the same procedures are followed as in training, except for the absence of data augmentation. 
Following the evaluation methodology used in DOTA \cite{DOTA}, the performance is measured in terms of PASCAL VOC2012 average precision (AP) \cite{VOC2012}, with IoU thresholds of 0.5 and 0.75. 
These metrics are referred to as AP50 and AP75, respectively. 
An SGD optimizer is employed with an initial learning rate of 0.01. 
Models are trained for a total of 12 epochs, with the learning rate decreasing by a factor of 10 at epochs 8 and 11. 
To ensure stable training, a linear warm-up strategy \cite{goyal2017accurate} is implemented for the first 500 iterations, using a learning ratio of 0.333. 
Gradient clipping is also applied with a maximum normalized value of 35 to prevent gradient explosion. 
The experiments are conducted using two NVIDIA 2080TI graphics processing units (GPUs) with a total batch size of 4. 
For testing, NMS with an Intersection over Union (IoU) threshold of 0.1 is utilized to remove duplicated bounding boxes. 
Additionally, boxes with scores lower than 0.05 are discarded to further reduce false detections.

\begin{table*}[htbp]
  \centering
  \setlength\tabcolsep{3pt}
  \caption{Performance comparison on CNBuilding-23P dataset in terms of AP50 on each province, and overall AP50 and AP75.}
    \resizebox{2.05\columnwidth}{!}{
    \begin{tabular}{cccccccccccccccccccccccccc}
    \toprule[1pt]
    Method & CQ    & FJ    & GS    & GD    & GX    & GZ    & HI    & HE    & HL    & HB    & HN    & JX    & JL    & LN    & OP    & NM    & QH    & SD    & SX    & SC    & XZ    & YN    & ZJ    & AP50  & AP75 \\
    \midrule
    Gliding Vertex \cite{GlidingVertex} & 51.8  & 51.2  & 56.9  & 52.8  & 63.1  & 45.3  & 62.0  & 63.1  & 65.0  & 49.0  & 65.8  & 67.3  & 72.2  & 47.9  & 49.7  & 60.9  & 65.1  & 58.5  & 56.0  & 44.5  & 63.2  & 62.8  & 58.0  & 54.6  & 25.6  \\
    KLD \cite{KLD} & 53.7  & 52.3  & 56.3  & 53.6  & 64.9  & 45.8  & 62.3  & 63.6  & 66.4  & 57.7  & 66.9  & 67.8  & 71.6  & 45.9  & 48.1  & 61.7  & 66.2  & 58.8  & 55.9  & 44.6  & 62.6  & 65.0  & 58.2  & 55.3  & 30.0  \\
    KFIoU \cite{KFIoU} & 50.6  & 51.8  & 56.2  & 52.9  & 62.4  & 43.4  & 61.5  & 60.6  & 64.1  & 57.8  & 65.7  & 66.6  & 72.1  & 44.9  & 47.2  & 60.8  & 65.3  & 57.7  & 53.4  & 43.2  & 61.7  & 63.8  & 56.8  & 53.9  & 26.0  \\
    FR-O \cite{FasterRCNN}  & 53.0  & 52.4  & 58.1  & 53.5  & 63.7  & 46.7  & 63.6  & 63.4  & 65.2  & 49.9  & 67.1  & 68.5  & 73.0  & 48.5  & 50.3  & 62.0  & 66.8  & 59.3  & 55.9  & 45.5  & 64.0  & 64.1  & 58.7  & 55.4  & 28.6  \\
    RoI-Transformer \cite{ROITrans} & 53.4  & 53.5  & 58.5  & 53.8  & 63.9  & 46.9  & 63.9  & 63.0  & 65.9  & 49.9  & 67.3  & 68.2  & 73.5  & 48.2  & 50.2  & 63.0  & 66.5  & 58.7  & 55.8  & 45.3  & 64.1  & 64.8  & 58.9  & 55.3  & 31.5  \\
    OR-CNN \cite{ORCNN} & 53.0  & 52.9  & 58.5  & 53.6  & 63.8  & 47.7  & 63.2  & 63.1  & 65.2  & 49.0  & 66.7  & 68.5  & 73.5  & 48.4  & 50.3  & 62.9  & 67.3  & 59.2  & 56.4  & 45.0  & 64.0  & 64.3  & 58.6  & 55.3  & 30.8  \\
    CR-O \cite{Cascade}  & 51.0  & 52.3  & 57.5  & 52.1  & 62.1  & 46.1  & 62.9  & 61.1  & 65.0  & 49.9  & 66.4  & 66.6  & 72.5  & 47.6  & 49.7  & 62.6  & 65.7  & 57.5  & 53.6  & 45.3  & 62.7  & 64.6  & 57.7  & 54.3  & 32.9  \\
    \midrule
    CEDet (Ours) & \textbf{54.9}  & \textbf{54.4}  & \textbf{59.5}  & \textbf{55.3}  & \textbf{64.8}  & \textbf{48.2}  & \textbf{65.3}  & \textbf{65.3}  & \textbf{67.6}  & \textbf{52.6}  & \textbf{68.3}  & \textbf{69.7}  & \textbf{75.0}  & \textbf{50.8}  & \textbf{52.4}  & \textbf{63.9}  & \textbf{69.1}  & \textbf{60.8}  & \textbf{58.7}  & \textbf{48.2}  & \textbf{65.3}  & \textbf{67.7}  & \textbf{61.1}  & \textbf{57.6}  & \textbf{33.3}  \\
    \bottomrule[1pt]
    \end{tabular}%
}
  \label{tab:CN_Results}%
\end{table*}%

\subsection{Comparison With Other Approaches}
This section compares the proposed CEDet with other detection methods on three building detection datasets: CNBuilding-9P, CNBuilding-23P, and SpaceNet. 
Table \ref{tab:CN9P_Results}, \ref{tab:CN_Results}, and \ref{tab:SpaceNet} detail the detection performance on different datasets. 

\textit{1) Results on CNBuilding-9P}: Table \ref{tab:CN9P_Results} presents the performance and inference speed of CEDet and other methods on CNBuilding-9P test set. 
Comparison methods include single-stage methods: Gliding Vertex \cite{GlidingVertex} (Gliding Vertex), KLD \cite{KLD} (KLD), and R3Det\cite{R3Det} with KFIoU \cite{KFIoU} (KFIoU); and multi-stage methods: Oriented Faster R-CNN \cite{FasterRCNN} (FR-O), RoI Transformer \cite{ROITrans} (RoI-Transformer), Oriented R-CNN \cite{ORCNN} (OR-CNN), and Oriented Cascade R-CNN (CR-O). 
Among them, CR-O is a variant that incorporates the structure of Cascade R-CNN \cite{Cascade} based on OR-CNN \cite{ORCNN}, accomplished by concatenating multiple OR-CNN Heads. 

Since CEDet introduces a cascade structure, CR-O and OR-CNN are selected jointly as the baseline for comparison. 
All models employ ResNet50 with FPN \cite{FPN} as the feature extractor and adopt the original configuration from MMRotate \cite{MMRotate}. 
Table \ref{tab:CN9P_Results} presents the AP50 results for each province, as well as the overall performance across all nine provinces. 
Furthermore, Table \ref{tab:CN9P_Results} includes the AP75 metric, which measures the high-precision detection ability of the model\cite{KLD}. 
The FPS column denotes the inference speed of the model when tested on a single GPU, encompassing the entire process of image processing, model inference, and post-processing operations. 

Table \ref{tab:CN9P_Results} illustrates that the performance of single-stage methods is typically inferior to that of multi-stage methods. 
Among the multi-stage methods, both RoI-Transformer \cite{ROITrans} and OR-CNN \cite{ORCNN} exhibit higher AP75 compared to FR-O \cite{Faster-RCNN}, suggesting that the utilization of oriented RoIs can enhance the precision of building detection.
However, the multi-stage methods exhibit similar performance in terms of AP50, indicating that these approaches possess similar classification capabilities on the CNBuilding-9P dataset.

\begin{figure*}[!htb]
	\centering
	\includegraphics[width=1\linewidth]{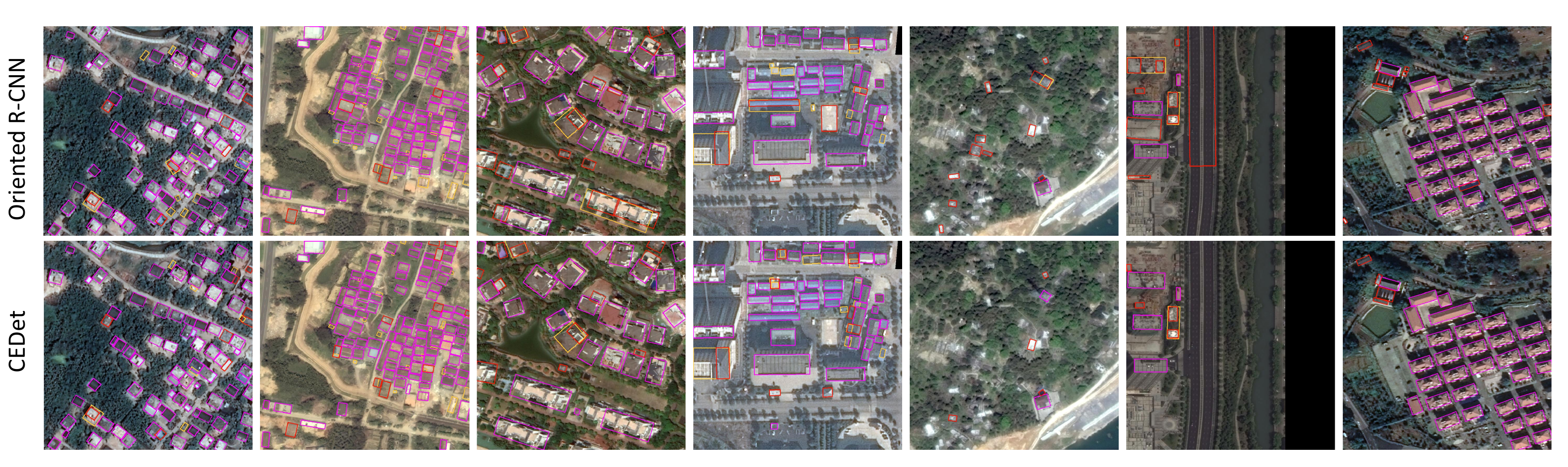}
	
	\caption{
 The visualization results of CEDet and OR-CNN \cite{ORCNN} on CNBuilding-9P dataset. 
 The pink boxes are the correct detections, defined as the IoU with the ground truth greater than 0.5. 
 Red boxes are false detection boxes. 
 Yellow boxes are ground truths that are not detected. }
\label{fig:Results}
\end{figure*}

Our CEDet achieves 58.5\% and 34.1\% in terms of AP50 and AP75, respectively. 
Compared with the baseline OR-CNN, our CEDet exhibits significant improvements across all provinces: +1.1\% on GS, +1.9\% on GD, +1.1\% on GX, +1.8\% on HI, +1.7\% on HN, +2.0\% on JL, +2.7\% on SD, +2.7\% on SC, and +2.7\% on ZJ. 
In summary, CEDet achieves an overall AP50 improvement of 2.1\% over OR-CNN and outperforms the best method, FR-O \cite{FasterRCNN}, by 1.7\%. 
This indicates that enhancing context can effectively improve the detector's ability to identify buildings in complex scenes. 
Additionally, due to the cascaded structure of CEDet, there is a 2.6\% improvement in AP75 compared to OR-CNN. 
While CR-O \cite{Cascade}, with its cascaded structure, performs closely in terms of AP75, it can be observed that CR-O's AP50 is approximately 1\% lower than other multi-stage methods. 
This demonstrates that the performance improvement of CEDet does not arise from the addition of detection stages but rather from the effective context extraction modules. 

Fig. \ref{fig:Results} presents the visualized detection results of CEDet and OR-CNN on CNBuilding-9P.  
Compared to OR-CNN, CEDet produces fewer false detection boxes and achieves more accurate building detection. 
CEDet incorporates an additional detection head and employs SGCM and ICMM to enhance features. 
As a result, CEDet exhibits the lowest inference speed among all detection methods, with a reduction of over half compared to OR-CNN. 
Subsequent ablation studies will comprehensively analyze the impact of each module on the detector's inference speed. 

\textit{2) Results on CNBuilding-23P}: 
The CNBuilding-23P dataset is approximately three times larger than the CNBuilding-9P dataset, providing a more comprehensive evaluation of the detector's detection ability in diverse and complex scenes. 
As shown in Table \ref{tab:CN_Results}, our CEDet achieves 57.6\% and 33.3\% in terms of AP50 and AP75, respectively. 
CEDet consistently outperforms OR-CNN in all provinces, showcasing its superiority. 
Notably, CEDet exhibits significant improvements in the provinces of HI (+2.1\%), HE (+2.2\%), HL (+2.4\%), HB (+3.6\%), LN (+2.4\%), OP (+2.1\%), SX (+2.3\%), SC (+3.2\%), YN (+3.4\%), and ZJ (+2.5\%). 
Furthermore, CEDet achieves a 2.3\% improvement over the baseline and a 2.2\% improvement over FR-O (55.4\%) in terms of overall AP50. 
These improvements demonstrate the effectiveness of CEDet in handling complex scenarios. 

\textit{2) Results on SpaceNet}\cite{van2018spacenet}: 
Table \ref{tab:SpaceNet} presents the detection performances on SpaceNet, measured in terms of AP50. 
Across all areas, multi-stage methods demonstrate higher accuracy compared to single-stage methods, with our CEDet achieving the best performance among all the methods considered. 
Compared with baseline OR-CNN \cite{ORCNN}, CEDet improves 2.3\% in the Khartoum area, 2.8\% in the Shanghai area, and 2.0\% in the Paris area. 
CEDet achieves a 2.4\% improvement over the baseline and a 1.1\% improvement over RoI-Transformer \cite{ROITrans} in terms of the overall AP50. 

\subsection{Ablation Study}
\textit{1) Effectiveness of SGCM}: SGCM plays a critical role in extracting multi-scale contextual information. 
By incorporating multi-scale feature fusion and reduction, each feature of FPN can access shared semantic features and acquire the ability to perceive contextual information at different scales \cite{LibraRCNN, HTC}. 
The inclusion of semantic segmentation loss directly contributes to the subsequent task by guiding the multi-scale features to develop explicit semantic recognition capabilities.
Furthermore, this multi-task paradigm proves advantageous in obtaining more generalized features \cite{HTC}. 

Table \ref{tab:SGCM} presents the performance improvement achieved by each module of SGCM. 
Additionally, we choose HTC \cite{HTC} and HSP \cite{HSP} as comparison methods. 
HTC removes the self-attention block within SGCM and increases the 2-layer $3\times 3$ convolutional layers to 4 layers.
On the other hand, HSP incorporates the ASPP module based on HTC to enhance the capability of capturing multi-scale features. 
In the fusion process, HSP replaces the 2-layer $3\times 3$ convolutional layers with the ASPP \cite{Context_ASPP}. 
The atrous rates in the ASPP are set to $(24, 48, 72)$ to be consistent with HSP \cite{HSP}.

\begin{table}[htbp]
  \centering
  \caption{Performance comparison on SpaceNet \cite{van2018spacenet} in terms of AP50 on each area, and overall AP50.
  }
    \setlength{\tabcolsep}{2.9mm}{
        \begin{tabular}{ccccc}
    \toprule[1pt]
    Method & Khartoum & Shanghai & Paris & AP50 \\
    \midrule
    Gliding Vertex \cite{GlidingVertex} & 50.88  & 56.26  & 60.44  & 55.86  \\
    KLD \cite{KLD}  & 49.89  & 55.67  & 65.37  & 56.98  \\
    KFIoU \cite{KFIoU} & 51.28  & 58.96  & 69.20  & 59.81  \\
    FR-O \cite{Faster-RCNN} & 53.05  & 58.38  & 69.44  & 60.29  \\
    RoI-Transformer \cite{ROITrans} & 53.35  & 59.22  & 71.77  & 61.45  \\
    OR-CNN \cite{ORCNN} & 52.01  & 58.26  & 70.25  & 60.17  \\
    CR-O \cite{Cascade} & 53.02  & 59.51  & 70.40  & 60.98  \\
    \midrule
    CEDet (Ours) & \textbf{54.36} & \textbf{61.07} & \textbf{72.25} & \textbf{62.56} \\
    \bottomrule[1pt]
    \end{tabular}%
    }
  \label{tab:SpaceNet}%
\end{table}%

\begin{table}[htbp]
  \centering
  \caption{Ablation studies about the SGCM module.}
      \setlength{\tabcolsep}{3mm}{
    \begin{tabular}{cccc}
    \toprule[1pt]
    Method & Context Enhancement & Semantic Loss & AP50 \\
    \midrule
    OR-CNN \cite{ORCNN} & -     & -     & 56.37  \\
    HTC \cite{HTC}   & -     & \checkmark     & 56.55  \\
    HSP\cite{HSP}   & ASPP \cite{Context_ASPP}  & \checkmark     & 56.44  \\
    SGCM  & Self-Attention \cite{ViT} & -     & 56.14  \\
    SGCM  & Self-Attention \cite{ViT}& \checkmark     & \textbf{56.90}  \\
    \bottomrule[1pt]
    \end{tabular}%
    }
  \label{tab:SGCM}%
\end{table}%

Compared with OR-CNN, using HTC results in a slight improvement of 0.18\% in AP50. 
However, there is a slight decrease in performance when employing ASPP to increase the receptive field. 
In contrast, SGCM's self-attention block improves HTC's performance by an additional 0.35\%, resulting in a total improvement of 0.53\% over OR-CNN. 
It is noticed that semantic segmentation loss is essential for extracting context features. 
A comparison of the bottom two rows in Table \ref{tab:SGCM} reveals that SGCM experiences a performance decrease of 0.76\% in the absence of semantic segmentation supervision. 

\begin{figure}[!htb]
	\centering
	\includegraphics[width=0.9\linewidth]{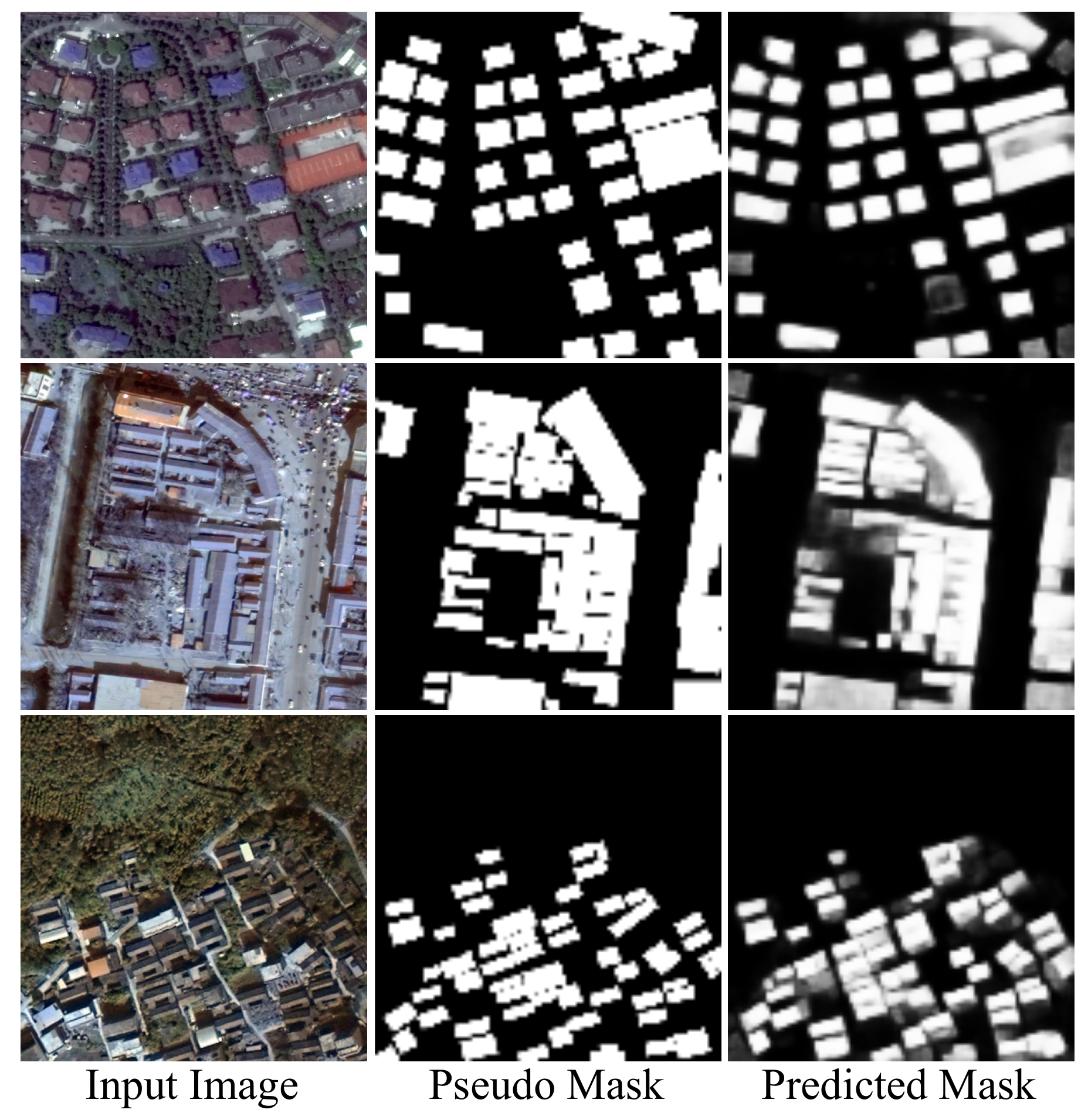}
	
	\caption{Visualization of the input images, target pseudo-masks, and the predicted masks.}
\label{fig:Semantic}
\end{figure}

Fig. \ref{fig:Semantic} illustrates the mask prediction obtained by SGCM. 
Despite the utilization of pseudo-labels for semantic segmentation, SGCM exhibits notable proficiency in accurately predicting the spatial extent of buildings, including those with intricately curved outlines, as exemplified in the second row.

\textit{2) Effectiveness of CE Head and ICMM}: 
Leveraging relational context to enhance building detection capabilities from complex remote sensing images has been identified as a viable approach \cite{Relation_CLT, Relation_HSDN}. 
In the proposed CEDet, CE Head incorporates ICMM to effectively capture relational contextual information via the spatial relationship between RoIs. 

\begin{table}[htbp]
  \centering
  \caption{Ablation studies about the CE Head}
      \setlength{\tabcolsep}{2.5mm}{
    \begin{tabular}{cccccc}
    \toprule[1pt]
     Experiment   & Head 1 & Head 2 & AP50  & AP75  & FPS \\
    \midrule
    OR-CNN \cite{ORCNN} & OR-CNN & -     & 56.4  & 31.5  & \textbf{18.6} \\
    Decoupled & D-OR-CNN & -     & 56.5  & 31.7  & 17.2 \\
    Replace & CE Head & -     & 57.7  & 31.7  & 12.9 \\
    Cascade & OR-CNN & CE Head & \textbf{58.2}  & \textbf{33.5}  & 11.1 \\
    \bottomrule[1pt]
    \end{tabular}%
    }
  \label{tab:CascadeHead}%
\end{table}%

Instead of replacing the OR-CNN Head, CEDet adds the CE Head after the OR-CNN Head for two reasons. 
Firstly, the oriented RoIs extracted by the oriented RPN often lack sufficient quality, which hinders the effective extraction of semantic features with high confidence. 
As a result, the CE Head struggles to learn effective instance relations. 
Secondly, adopting the cascade structure has proven to enhance detection accuracy, and the joint decision-making of multi-stage classification scores further improves classification accuracy \cite{Cascade}. 
Additionally, to mitigate the feature ambiguity caused by the aggregation of contextual information, CE Head employs a decoupling structure \cite{wu2020rethinking} and exclusively utilizes ICMM in the classification branch, thereby minimizing its impact on regression.

Table \ref{tab:CascadeHead} compares the effects of these different designs. 
Replacing OR-CNN Head with CE Head results in a 1.3\% improvement in AP50, while the improvement in AP75 is marginal. 
Subsequently, introducing the cascade structure leads to a further 0.5\% improvement in AP50. 
Notably, the AP75 metric demonstrates significant improvement, surpassing OR-CNN by 2\%.

The second row of Table \ref{tab:CascadeHead} presents the impact of replacing the original OR-CNN with the decoupling structure. 
In the original OR-CNN Head, RoI features are extracted using two fully connected layers, which are then utilized for classification and regression. 
Conversely, the decoupled OR-CNN Head utilizes four independent fully connected layers to extract RoI features and performs classification and regression separately. 
However, experiments demonstrate that the decoupled head fails to improve the detector's performance effectively. 
This suggests that the performance gain achieved by CE Head is not attributed to parameter increments but rather to the utilization of relational context. 

\begin{table}[htbp]
  \centering
  \caption{Ablation studies about the effectiveness of NMS in ICMM}
    \setlength{\tabcolsep}{5.5mm}{
    \begin{tabular}{ccccc}
    \toprule[1pt]
     Method  & NMS   & AP50  & AP75  & FPS \\
    \midrule
    ICMM  &   -  & 57.99  & 33.06  & \textbf{11.20}  \\
    ICMM  & \checkmark     & \textbf{58.24}  & \textbf{33.50}  & 11.10  \\
    \bottomrule[1pt]
\end{tabular}%
}
  \label{tab:NMS}%
\end{table}%

In addition to the cascade structure, NMS is introduced to suppress noisy oriented RoIs. 
Table \ref{tab:NMS} analyses this strategy. 
The results reveal a 0.25\% improvement in AP50 and a 0.5\% improvement in AP75 when NMS is applied.
These findings suggest that removing noisy oriented RoIs facilitates ICMM in capturing relationship information more effectively.

\begin{table}[htbp]
  \centering
  \caption{Ablation studies about the number of ICMMs}
    \setlength{\tabcolsep}{3.5mm}{
    \begin{tabular}{cccccc}
    \toprule[1pt]
     \# of ICMMs  & 1 & 2 & 3 & 4 & 5 \\
    \midrule
    AP50  & 58.17  & \textbf{58.24}  & 58.07  & 58.16  & 58.20  \\
    AP75  & 33.41  & 33.50  & \textbf{33.52}  & \textbf{33.52}  & 33.43  \\
    FPS   & \textbf{11.2}  & 11.1  & 11    & 11    & 10.8 \\
    \bottomrule[1pt]
    \end{tabular}%
    }
  \label{tab:GCNLayer}%
\end{table}%

ICMM can be stacked multiple times to obtain more decisive relational contextual information. 
The impact of the number of ICMMs is examined in Table \ref{tab:GCNLayer}. 
The results indicate that the performance is not significantly affected by the number of ICMMs. 
Even with a single ICMM, there is an improvement in performance. 
Increasing the number of ICMMs does not lead to better performance but affects the inference speed. 
This observation can be attributed to the relatively straightforward relational context in the building dataset, where a simpler design is sufficient to extract the relational context effectively. 
As a result, the CE Head in CEDet employs only two ICMMs.

\begin{table}[htbp]
  \centering
  \caption{Ablation studies about the impact of modular combinations}
    \setlength{\tabcolsep}{2.5mm}{
    \begin{tabular}{ccccccc}
    \toprule[1pt]
    Method & Cascade  & SGCM  & ICMM  & AP50 & AP75 & FPS \\
    \midrule
    Baseline &    -  &   -  &    -  & 56.4  & 31.5  & \textbf{18.6}  \\
    \midrule
    \multirow{5}[1]{*}{CEDet} & \checkmark     &       &       & 55.6  & 33.9  & 11.5  \\
          &   -   & \checkmark     &   -    & 56.9  & 31.6  & 16.2  \\
          &   -   &   -    & \checkmark     & 57.7  & 31.7  & 12.9  \\
          & \checkmark     &   -    & \checkmark     & 58.2  & 33.5  & 11.1  \\
          & \checkmark     & \checkmark     & \checkmark     & \textbf{58.5}  & \textbf{34.1}  & 8.9  \\
    \bottomrule[1pt]
    \end{tabular}%
    }
  \label{tab:Combine}%
\end{table}%

\begin{table*}[htbp]
  \centering
  \caption{Ablation studies about different backbone architectures. 
  ResNet-50 \cite{ResNet}, ResNet-101 \cite{ResNet}, ResNext-101 \cite{ResNext}, and Swin-T \cite{Swin} are adopted in the experiment}
    \setlength{\tabcolsep}{3.6mm}{
    \begin{tabular}{c|cccccccccccc}
    \toprule[1pt]
    Method & Backbone & GS    & GD    & GX    & HI    & HN    & JL    & SD    & SC    & ZJ    & AP50  & AP75 \\
    \midrule
    \multirow{4}[1]{*}{OR-CNN \cite{ORCNN}} & ResNet-50  & 56.0  & 52.0  & 62.9  & 62.0  & 64.6  & 72.4  & 58.6  & 40.1  & 55.1  & 56.4  & 31.5  \\
          & ResNet-101  & 55.6  & 51.7  & 62.9  & 61.4  & 63.7  & 71.7  & 58.7  & 41.5  & 55.6  & 56.4  & 31.8  \\
          & ResNext-101 & 56.7  & 52.0  & 63.5  & 62.2  & 64.7  & 72.1  & 60.2  & 42.5  & 58.6  & 57.6  & 32.8  \\
          & Swin-T & 55.4  & 51.1  & 62.1  & 60.0  & 64.4  & 71.9  & 57.6  & 39.7  & 55.0  & 55.7  & 30.1  \\
    \midrule
    \multirow{4}[1]{*}{CEDet} & ResNet-50  & 57.1  & 53.9  & 64.0  & 63.8  & 66.3  & 74.4  & 61.3  & 42.8  & 57.8  & 58.5  & 34.1  \\
          & ResNet-101 & 56.3  & 52.9  & 62.9  & 62.6  & 64.5  & 73.9  & 59.8  & 43.5  & 58.2  & 57.9  & 33.6  \\
          & ResNext-101 & 57.5  & 53.5  & 64.0  & 63.4  & 66.2  & 73.9  & 61.2  & \textbf{44.8}  & 59.7  & 59.0  & \textbf{34.6}  \\
          & Swin-T & \textbf{58.7} & \textbf{53.9} & \textbf{64.8} & \textbf{64.0} & \textbf{67.9} & \textbf{75.3} & \textbf{62.2} & 44.3 & \textbf{59.9} & \textbf{59.6} & 34.2 \\
    \bottomrule[1pt]
    \end{tabular}%
    }
  \label{tab:Backbone}%
\end{table*}%

\textit{3) Combination of Modules}: 
Table \ref{tab:Combine} presents a comprehensive overview of the performance enhancements achieved by integrating various modules into the baseline OR-CNN \cite{ORCNN}. 
The cascade structure \cite{Cascade} demonstrates effective performance improvement at AP75 (+2.4\%). 
However, it does not enhance building classification ability, as evidenced by a slight decrease of 0.8\% in AP50. 
Without sufficient exploitation of contextual information, the detector's recognition capability in complex scenes cannot be effectively improved. 
SGCM improves AP50 by 0.5\% through multi-scale context mining. 
On the other hand, ICMM achieves a more significant improvement of 1.3\% in AP50 through relation-based context mining. 
This can be attributed to ICMM's utilization of high-level semantic features for reasoning and its ability to model relations with rotation and scale invariance, facilitating deeper contextual information extraction. 
The combination of ICMM and the cascade structure further enhances performance, resulting in a 1.8\% improvement over OR-CNN in AP50. 
Finally, the proposed CEDet with ICMM and SGCM leads to a remarkable 2.1\% improvement compared with the baseline. 

\textit{4) Different Backbones}: 
A stronger backbone can lead to better generalization, thereby enhancing the recognition ability of the detector in complex real-world scenes. 
In the case of CEDet, features with enhanced generalization also contribute to the improved extraction capability of contextual information. 
Table \ref{tab:Backbone} presents a performance comparison between OR-CNN \cite{ORCNN} and CEDet under different backbone architectures on CNBuilding-9P dataset, including ResNet50 \cite{ResNet}, ResNet101 \cite{ResNet}, Swin-T \cite{Swin}, and ResNext101 \cite{ResNext}. 
For ResNext101, OR-CNN achieves a 1.2\% improvement over the baseline in AP50 compared with ResNet50. However, using Swin-T as the backbone results in a decrease of 0.7\% in AP50. 
It is worth noting that OR-CNN with different backbone architectures does not outperform CEDet with ResNet50. 
On the other hand, CEDet with Swin-T backbone leads to a further improvement of 0.9\% in AP50 compared with CEDet with ResNet50 backbone, which highlights the effectiveness of Swin-T backbone in enhancing the performance of CEDet.

\section{Conclusion}
This paper introduces CEDet, a novel approach that effectively leverages contextual information to achieve highly accurate building detection. 
CEDet adopts a multi-stage structure to enhance the ability of contextual feature extraction. 
The proposed SGCM module in CEDet addresses the issue of insufficient long-range feature interactions observed in existing multi-scale context extraction methods by utilizing a self-attention mechanism.
A semantic segmentation loss based on pseudo-masks is also employed to supervise contextual feature extraction. 
The Instance Context Mining Module (ICMM) is proposed to capture contextual information between instances through spatial relationships, significantly enhancing the detector's accuracy. 
Finally, ablation experiments demonstrate the effectiveness of SGCM and ICMM. 
Moreover, our CEDet achieves outstanding performance on three benchmark datasets, i.e., CNBuilding-9P, CNBuilding-23P, and SpaceNet, further illustrating its superiority.




\bibliographystyle{IEEEtran}
\bibliography{paper}

\end{document}